\documentclass[]{fairmeta}
\usepackage{setspace}

\usepackage{amsmath,amsfonts,amsthm,amssymb,bm}

\def\eqref#1{equation~\ref{#1}}

\def\1{\bm{1}}

\DeclareMathAlphabet{\mathsfit}{\encodingdefault}{\sfdefault}{m}{sl}
\SetMathAlphabet{\mathsfit}{bold}{\encodingdefault}{\sfdefault}{bx}{n}

\newcommand{\E}{\mathbb{E}}

\theoremstyle{definition}

\theoremstyle{definition}

\usepackage{style}

\usepackage{times}
\usepackage{latexsym}
\usepackage{textcomp}
\usepackage{cuted}

\usepackage[T1]{fontenc}

\usepackage[utf8]{inputenc}

\usepackage{microtype}

\usepackage{inconsolata}

\usepackage[]{mdframed}

\definecolor{orange}{rgb}{1,0.5,0}
\definecolor{mdred}{rgb}{0.7,0,0}
\definecolor{mdgreen}{rgb}{0.05,0.6,0.05}
\definecolor{mdblue}{rgb}{0,0,0.7}
\definecolor{dkblue}{rgb}{0,0,0.5}
\definecolor{dkgreen}{rgb}{0,0.5,0}
\definecolor{dkgray}{rgb}{0.3,0.3,0.3}
\definecolor{slate}{rgb}{0.25,0.25,0.4}
\definecolor{gray}{rgb}{0.5,0.5,0.5}
\definecolor{ltgray}{rgb}{0.7,0.7,0.7}
\definecolor{purple}{rgb}{0.7,0,1.0}
\definecolor{lavender}{rgb}{0.65,0.55,1.0}
\definecolor{theme}{HTML}{6b8a97}
\definecolor{unchanged}{rgb}{0.7,0.7,0.7}
\definecolor{emphasisbg}{rgb}{0.9,0.9,0.9}

\newcommand{\zerodisplayskips}{%
  \setlength{\abovedisplayskip}{7pt}%
  \setlength{\belowdisplayskip}{7pt}%
  \setlength{\abovedisplayshortskip}{7pt}%
  \setlength{\belowdisplayshortskip}{7pt}}
\appto{\normalsize}{\zerodisplayskips}
\appto{\small}{\zerodisplayskips}
\appto{\footnotesize}{\zerodisplayskips}

\newif\iflongversion
\longversiontrue

\newif\ifonecolumn
\onecolumntrue

\title{Parallel-SFT: Improving Zero-Shot Cross-Programming-Language Transfer for Code RL}

\author[\text{\Aquarius,\Cancer}]{Zhaofeng Wu}
\author[\text{\Aquarius}]{Shiqi Wang}
\author[\text{\Aquarius}]{Emma Peng}
\author[\text{\Aquarius}]{Anuj Goyal}
\author[\text{\Aquarius}]{Melanie Kambadur}
\author[\text{\Aquarius}]{Sebastian Ruder}
\author[\text{\Cancer}]{Yoon Kim}
\author[\text{\Aquarius}]{Chloe Bi}

\affiliation[\text{\Aquarius}]{Meta Superintelligence Labs}
\affiliation[\text{\Cancer}]{MIT}

\contribution[*]{Work done during ZW's internship at Meta}

\abstract{Modern language models demonstrate impressive coding capabilities in common programming languages (PLs), such as C++ and Python, but their performance in lower-resource PLs is often limited by training data availability. In principle, however, most programming skills are universal across PLs, so the capability acquired in one PL should transfer to others.
In this work, we propose the task of zero-shot cross-programming-language transfer for code RL. We find that, for Llama-3.1, RL training for code generation in a source PL fails to improve, and sometimes even degrades, the performance on other target PLs.
To address this, we hypothesize that effective RL transfer requires a generalizable SFT initialization before RL.
We thus propose \textbf{Parallel-SFT}, an SFT strategy that incorporates ``parallel programs''---functionally equivalent code implemented in multiple PLs---into the data mixture.
We demonstrate that this improves transferability: when we subsequently perform RL on our Parallel-SFT model, we observe better generalization to unseen PLs.
Analysis of the model internal representations reveals that Parallel-SFT leads to a more functionality-centric latent space, where equivalent programs across PLs are more tightly clustered, which we hypothesize to contribute to the improved transferability.
}

\date{\today}

\begin{document}

\maketitle

\abstract{Modern language models demonstrate impressive coding capabilities in common programming languages (PLs), such as C++ and Python, but their performance in lower-resource PLs is often limited by training data availability. In principle, however, most programming skills are universal across PLs, so the capability acquired in one PL should transfer to others.
In this work, we propose the task of zero-shot cross-programming-language transfer for code RL. We find that, for Llama-3.1, RL training for code generation in a source PL fails to improve, and sometimes even degrades, the performance on other target PLs.
To address this, we hypothesize that effective RL transfer requires a generalizable SFT initialization before RL.
We thus propose \textbf{Parallel-SFT}, an SFT strategy that incorporates ``parallel programs''---functionally equivalent code implemented in multiple PLs---into the data mixture.
We demonstrate that this improves transferability: when we subsequently perform RL on our Parallel-SFT model, we observe better generalization to unseen PLs.
Analysis of the model internal representations reveals that Parallel-SFT leads to a more functionality-centric latent space, where equivalent programs across PLs are more tightly clustered, which we hypothesize to contribute to the improved transferability.
}

\ifonecolumn
\begin{figure}[h!]
    \centering
    \includegraphics[width=0.41\columnwidth]{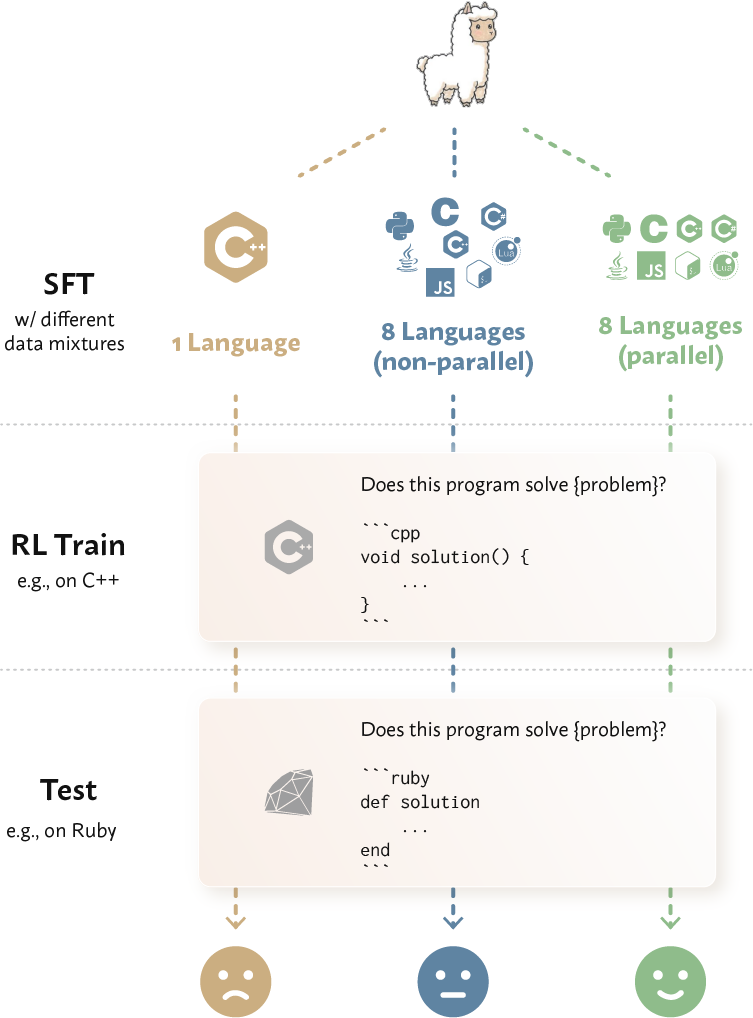}
    \caption{We propose zero-shot cross-PL transfer for code RL, where we perform RL training and testing on different PLs. Llama-3.1 fails to transfer effectively~(\S\ref{sec:eval}; not represented in this figure). We find that, compared to single-PL SFT, multi-PL SFT improves transferability in downstream RL, which is further enhanced when the SFT code programs are parallel~(\S\ref{sec:method}). The code validation task is shown here as an example.}
    \label{fig:overview}
\end{figure}
\else
\begin{figure}[t!]
    \centering
    \includegraphics[width=\linewidth]{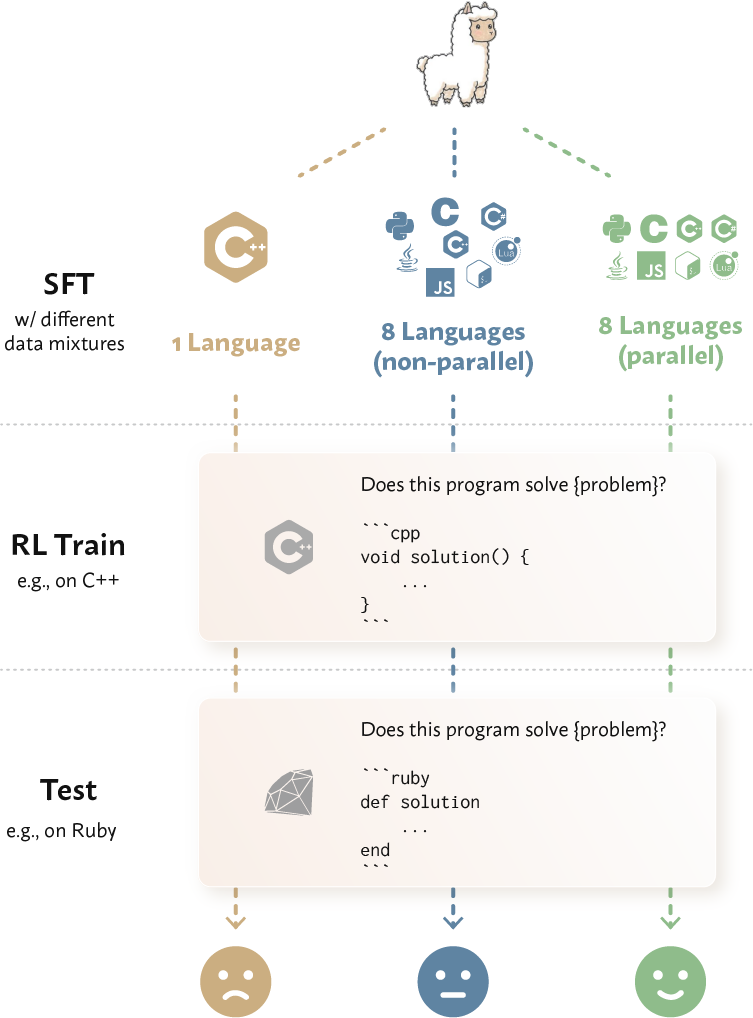}
    \caption{We propose zero-shot cross-PL transfer for code RL, where we perform RL training and testing on different PLs. Llama-3.1 fails to transfer effectively~(\S\ref{sec:eval}; not represented in this figure). We find that, compared to single-PL SFT, multi-PL SFT improves transferability in downstream RL, which is further enhanced when the SFT code programs are parallel~(\S\ref{sec:method}). The code validation task is shown here as an example.}
    \label{fig:overview}
\end{figure}
\fi

\section{Introduction}

As language models (LMs) are increasingly integrated into real-world coding workflows, their ability to handle the long tail of programming languages (PLs) becomes critical. Yet, LMs often show vastly disparate performance across PLs depending on their resource availability~\citep{multiple,multiplt,10.1145/3689728,xu-etal-2025-cruxeval,chai2025mceval}.
This disparity signals a fundamental inefficiency in leveraging universal code reasoning.
Analogous to human programmers who can learn algorithms in a certain source PL and reproduce them in other target PLs, LMs ideally should also generalize their programming capabilities across PLs.

In this work, we focus on this PL generalization problem, specifically within the RL stage of code post-training, where task-specific training usually takes place.
We define the task of \textbf{zero-shot cross-programming-language transfer}, optimizing a task policy via RL in a source PL and zero-shot evaluating its performance on another target PL.
Empirically, we find that Llama-3.1~\citep{grattafiori2024llama3herdmodels} exhibits limited transferability: while RL yields consistent benefits within the source PL, these gains fail to transfer to different target PLs, sometimes even degrading the performance.

To overcome this limitation, we hypothesize that successful RL transfer requires a generalizable SFT initialization.
We propose \textbf{Parallel-SFT}, which incorporates synthetically generated ``parallel programs''---functionally-equivalent programs across PLs---into the SFT data mixture (Figure~\ref{fig:overview}).
Rather than merely modeling the surface form of code tokens, Parallel-SFT semantically grounds programs with execution equivalence.
Across two coding tasks, 
Parallel-SFT outperforms both a single-source-PL baseline and a standard multi-PL baseline (without parallel programs) in downstream RL transfer.
Remarkably, Parallel-SFT sometimes even outperforms direct training on the target PL, surpassing the standard ``oracle'' data-rich setting, highlighting the effectiveness of our method.

In the multilingual NLP literature, successful LM generalizability between natural languages is often attributed to language-general representations, where multilingual models develop representations that capture the underlying semantics of texts~\citepia{pires-etal-2019-multilingual,wu-dredze-2019-beto,conneau-etal-2020-emerging}.
Similarly, we hypothesize that Parallel-SFT aligns the PL representation space, leading to more semantic code representations.
This enables the policy improvements learned during source-PL RL to better propagate to target PLs.
Our analysis supports this hypothesis: Parallel-SFT induces higher similarity between parallel programs even in \emph{unseen} PLs.
We hope our method inspires future work on training generalizable models not only across PLs, but also across natural languages, domains, and tasks.

\section{Background}

In this section, we outline the coding tasks considered in this work and the standard post-training pipeline to solve them.
We then review the paradigm of zero-shot cross-lingual transfer in NLP, which motivates our experimental setup.

\subsection{Coding Tasks} \label{sec:coding-tasks}

We focus on two coding tasks: code generation and code validation.

In \textbf{code generation}, the model takes a natural language description of a coding question, $q$, and generates a solution code program $c$. $c$ is evaluated against a suite of instance-specific test cases, and is considered correct iff it passes all. Performance is typically measured using the pass@$k$ metric: for each question, the model samples $k$ solutions, and the instance is considered solved if at least one sample passes all tests.
Broadly, the field distinguishes between two difficulty levels: competition coding that consists of complex Olympiad-level questions~\citep{Li_2022}, and more general coding questions that are familiar to software engineers~\citepia{chen2021evaluatinglargelanguagemodels,austin2021programsynthesislargelanguage,hendrycks2021measuring}. In this work, we evaluate on the former, but train on the latter too.

In \textbf{code validation}, the model takes as input a tuple $(q, c)$ containing a question and a candidate code solution, and it outputs whether $c$ is correct, i.e., whether $c$ solves $q$. Performance is measured by boolean accuracy.\footnote{Code validation models are useful in RL post-training as reward models to estimate the correctness of generated programs, circumventing the need for tests and the sometimes high execution cost.}

\subsection{Training for Coding Tasks}

The standard recipe for training coding tasks includes three stages: pretraining, supervised finetuning (SFT), and RL. First, a model is pretrained on massive corpora of text and code. 
Second, the model undergoes SFT on a dataset of instruction and ground-truth response pairs, $(x^{\text{SFT}}, y^{\text{SFT}})$.
Like in pretraining, the SFT data also usually includes both natural language tasks and coding tasks.

Finally, we perform RL to optimize for a specific coding task. It requires a dataset of prompts $x^{\text{RL}}$ and a verifier.\footnote{Or a reward model for non-verifiable tasks, which we do not consider.} For code generation, $x^{\text{RL}}=q$, the coding question, and the verifier is a code executor with test cases. For code validation, $x^{\text{RL}}=(q,c)$,\footnote{Precisely, the $x^{\text{RL}}$ in both cases also requires a template.} and the verifier extracts the boolean prediction from the code solution and compares it against the ground truth.
During training, the policy model samples a response, which is scored by the verifier. This binary signal is used to update the policy using RL algorithms such as GRPO~\citep{shao2024deepseekmathpushinglimitsmathematical}.

\begin{figure*}[t!]
    \centering
    \includegraphics[width=\textwidth]{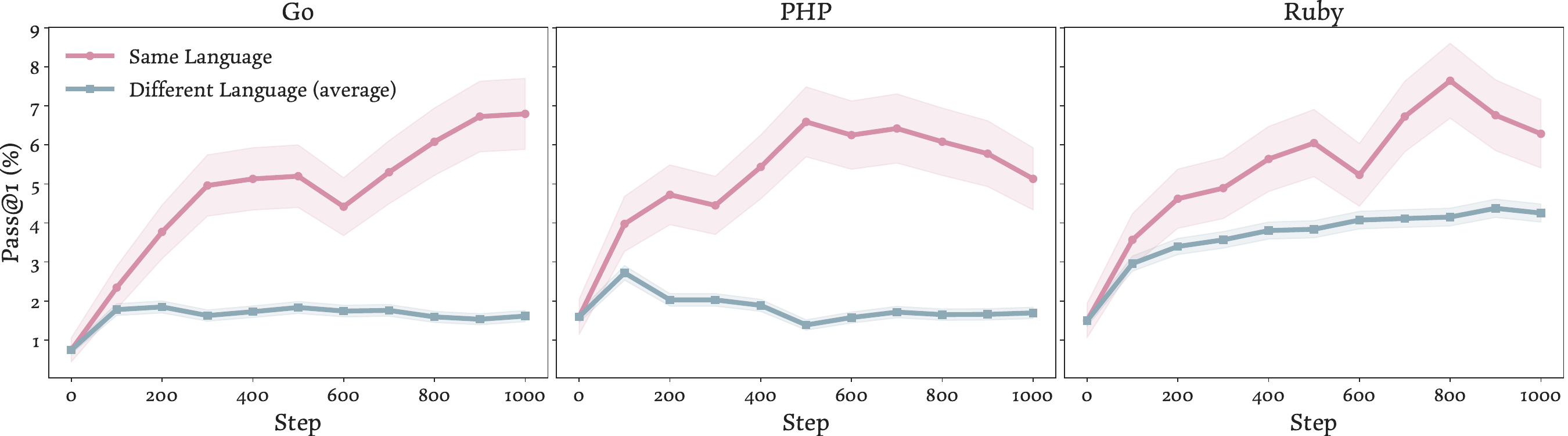}
    \vspace{-6mm}
    \caption{\textbf{Failure of Naive Cross-Lingual Transfer.} We visualize the pass@1 performance on a target language (e.g., Go) during RL training on the same vs. different source languages (averaged over 10 source PLs). Shaded regions indicate 95\% confidence intervals. While RL on the target language itself leads to consistent gains, cross-PL transfer from a different source language leads to limited improvements or even performance degradation.}
    \label{fig:eval}
\end{figure*}

\subsection{Zero-Shot Cross-Lingual Transfer} \label{sec:bg-nl}

Zero-shot cross-lingual transfer is a core paradigm in multilingual NLP.
In this setup, a multilingual model is finetuned on task data in a \textit{source} language and then evaluated zero-shot on the same task in a different \textit{target} language.
Despite never seeing any task data in the target language, models often achieve surprisingly high performance~\citep{huang-etal-2019-unicoder,xlm,conneau-etal-2020-unsupervised}, provided that they were exposed to unsupervised target-language data during pretraining.
This transferability is often attributed to language-general representations~\citep{conneau-etal-2020-emerging}.
This is most intuitive when an LM is finetuned by training a task-specific head while freezing the pretrained parameters~\citepia{peters-etal-2018-deep,peters-etal-2019-tune,artetxe-schwenk-2019-massively}, where language-general representations allow the task head's decision boundary to naturally generalize across languages~\citep{conneau-etal-2018-xnli}.
While modern models typically update the pretrained parameters too, a language-general initialization still enables a general learning process.

While this paradigm is successful for natural languages, its application to programming languages has yielded mixed results. Early work focused on pretraining encoder models on code-only corpora~\citep{feng-etal-2020-codebert,wang-etal-2021-codet5}.
These studies typically involved non-generative tasks (e.g., code search; \citealp{10.1145/3524610.3527917,baltaji2025crosslingual})
or training a separate decoder from scratch to pair with the pretrained encoder~\citep{Ahmed2021MultilingualTF,10.1145/3524610.3527917}.
Despite mostly positive transfer effects, these settings are far removed from the current pretraining paradigm.
Recently, \citet{baltaji2025crosslingual} investigated cross-PL code generation in modern LLMs via in-context learning, but found that cross-PL transfer often underperforms zero-shot prompting.
Furthermore, interpretability studies have revealed low cross-PL representation generality in code models~\citep{utpala-etal-2024-language}, potentially limiting transferability.

\section{Cross-Programming-Language Generalization Evaluation} \label{sec:eval}

We quantify the zero-shot cross-lingual transferability of current models for code RL.
Starting from an LM, we perform RL for a coding task in a \emph{source} PL, and then zero-shot evaluate the resulting model on the same task in another \emph{target} PL.

\paragraph{Model \& Task.}
We use Llama-3.1-8B-Instruct \citep{grattafiori2024llama3herdmodels} as our initial model.\footnote{It is not an SFT model, which typically precedes the RL stage, but it has also gone through preference tuning with DPO. We do not believe this affects our results, especially since it is common for current LMs to undergo multiple stages of policy optimization~\citepia{qwen3,bercovich2025llamanemotronefficientreasoningmodels}. In \S\ref{sec:eval-custom-sft}, we validate this with our own custom-trained SFT models and observe qualitatively similar results.}\footnote{While Qwen models~\citep{qwen25,qwen3} have higher raw coding scores, past work has suggested likely substantial data contamination on coding benchmarks, and Llama models to a much less extent~\citep{wu2025reasoningmemorizationunreliableresults}. We thus choose to use Llama-3.1 as a cleaner testbed.\label{fn:why-not-qwen}} We consider the task of competition-level code generation and use the CodeForces dataset for training and evaluation~\citep{penedo2025codeforces}.
See dataset statistics in \S\ref{sec:rl-dataset-stats}.

\paragraph{Language Selection.}
Since our goal is to improve performance on medium- and low-resource PLs, we select 3 diverse target PLs: Go, PHP, and Ruby. Respectively, they rank 15th, 16th, and 25th on the TIOBE popularity index (as of December 2025), even lower than PLs like Ada and MATLAB~\footnote{\url{https://www.tiobe.com/tiobe-index/}}. This categorization aligns with past work on medium-/low-resource code modeling~\citep{10.1145/3510003.3510049,10.1609/aaai.v38i16.29749,multiple}. We also consider 8 additional languages that we only treat as source: Python, C, C++, Java, C\#, JavaScript, Bash, and Lua.

\paragraph{Results.}
Figure~\ref{fig:eval} shows the transfer performance throughout RL training, averaged over 10 source languages (11 PLs in total, minus the target language itself). We also show the in-PL performance, where the same target language is used for RL training.
We see that while training using the target language yields consistent improvements, transferring from another source language leads to limited, or sometimes degraded effects (negative transfer).
This is consistent with the observation from \citet{baltaji2025crosslingual} that was based on in-context learning rather than RL.
In \S\ref{sec:eval-custom-sft}, we show similar results initializing not from the instruct model but the base pretrained model and performing our own source-PL SFT. This confirms that this is a fundamental issue and not due to model variations.

\section{Method: Parallel-SFT} \label{sec:method}

Observing unsatisfactory cross-PL generalization, in contrast to prior success for natural languages (\S\ref{sec:bg-nl}), we reflect on differences between natural languages and programming languages.
Past work on multilingual NLP has attributed multilingual capabilities to organic parallel texts (i.e., translations) in pretraining corpora.
They function like a ``Rosetta Stone'' to help align the model representations of different languages.
\citet{info16121077} distinguishes between two kinds of such parallel texts, ``local'' (parallel texts within a single pretraining instance) and ``global'' (parallel texts across different pretraining instances).
\textbf{We argue that both signals are lacking for programming languages.}

First, natural language corpora benefit from ``incidental bilingualism,'' where a single document contains parallel text.
For example, \citet{briakou-etal-2023-searching} found that 0.34\% of PaLM's~\citep{chowdhery2022palmscalinglanguagemodeling} pretraining data contains parallel sentences, totaling over 30 million pairs.
In contrast, code files are almost exclusively monolingual, let alone containing parallel programs.

Globally, across pretraining instances, parallel sentences abound in training corpora: people discuss similar topics in all languages.
Fundamentally, this is because natural languages are grounded in the same physical world and model the same ``underlying reality''~\citep{huh2024platonic}.
This creates a shared semantic structure that models leverage, for example as their generalizability benefits from domain similarity~\citep{lample2018word,conneau-etal-2020-emerging}.
Programming languages, however, are often domain-specialized. It is rare for an LM training script in Python to have a direct counterpart in Ruby or C++, as those languages are typically applied to different domains (e.g., web development or systems programming).
This domain mismatch limits the organic emergence of parallel signals, and in turn leads to empirically fragmented representations across PLs~\citep{utpala-etal-2024-language} and weak semantic understanding in lower-resource PLs~\citep{gu2025tasks}.
This contributes to the negative transfer effects observed in \S\ref{sec:eval}.

To bridge this gap,
we propose \textbf{Parallel-SFT}: a method to improve cross-PL capabilities by explicitly constructing ``parallel programs''---functionally equivalent code programs across PLs---anchored by (near-)identical natural language instructions.
This is reminiscent of second-order co-occurrences in word models, where models can infer from a sentence pair ``I have an adorable pet \underline{Corgi}.'' and ``I have an adorable pet \underline{Samoyed}.'' the relationship between the words ``\underline{Corgi}'' and ``\underline{Samoyed}''.
Similarly, we hypothesize we can improve the model's cross-PL representation with training instances such as ``Write a Python implementation of binary search. \underline{\texttt{def binary\_search(...}}'' and ``Write a Java implementation of binary search. \underline{\texttt{public int binarySearch(...}}''.
Intuitively, this semantically grounds code models with execution equivalence.

Formally, our SFT data follows the format $\left\{\left((q_i, c_i^{\text{Python}}), (q_i, c_i^{\text{C++}}), \cdots\right)\right\}_i$, where the $c^{\text{lang}}$ are mutual translations that all solve $q$.
By forcing the model to map the same question $q$ to different language-specific realizations $c^{\text{lang}}$, Parallel-SFT encourages a functionality-centric, rather than syntax-centric, representation space.
We will show that Parallel-SFT enables a more generalizable SFT model: the subsequent RL stage in a source language transfers better to target languages.

\section{Experimental Setup}

\subsection{Parallel Data Construction} \label{sec:parallel-data-construction}

We construct a high-quality parallel program dataset.
Following our examples in \S\ref{sec:method} and common practices~\citepia{roziere2024codellamaopenfoundation,lambert2025tulu3pushingfrontiers}, we use instances of the \emph{code generation} task for SFT training.
We utilize
a combination of the APPS~\citep{hendrycks2021measuring} and CodeContests~\citep{codecontests} datasets, which contain diverse code generation instances.
To ensure rigorous evaluation, we filter the data to exclude any CodeForces data which we reserve for RL training and evaluation (\S\ref{sec:setup-rl}).
We also filter out questions that have visual inputs.

These datasets contain mostly solutions in Python.
To generate parallel programs in other languages, we perform synthetic code translation:
for each instance $(q, c^{\text{Python}})$, we translate $c^{\text{Python}}$ into $c^{\text{C++}}$, $c^{\text{Java}}$, etc.,
by prompting Llama-4-Maverick, a 400B-parameter mixture-of-expert model~\citep{meta2025llama4}. We include our translation prompt in \S\ref{sec:translation-prompt}. This model generally has good translation accuracy: it achieves 57.3\% pass@1 when validated on the tests in
the datasets.
We further filter out all translations that fail the tests to ensure translation quality.\footnote{As a sanity check, we also filter out original Python solutions that fail the tests in our execution environment.}

These datasets contain
multiple Python solutions per question. We sample 8 generations per source $c^{\text{Python}}$ solution to ensure an abundance of candidate programs. Our final synthetic dataset contains 3,111 unique questions and, on average, 181 instances per language per question, totaling $\approx$6.2 million instances.
Formally, the dataset is
$\{ (q_i, (C_{i}^{\text{Python}}, C_{i}^{\text{C++}}, \cdots ) ) \}_{i=1}^{N}$
where $N=3,111$. Each set $C_{i}^{\text{lang}} = \{c_{ij}^{\text{lang}}\}_{j=1}^{N_i^{\text{lang}}}$ is the set of verified code solutions in PL ``$\text{lang}$'', with $N_i^{\text{lang}}$ averaging to 181.
This amplifies the parallel program signal: if a question is only seen once per language, it could easily be forgotten in the course of training.

\subsection{SFT Dataset Mixtures} \label{sec:sft-dataset}

We construct a realistic general-purpose SFT dataset by utilizing the Tulu 3 SFT mixture (939k instances; \citealp{lambert2025tulu3pushingfrontiers}).
We only retain its natural language data on instruction following, safety, etc.
To be clean, we remove its coding instances (N=142k) and replace them with an equal-sized sample from our synthetic dataset in \S\ref{sec:parallel-data-construction}.

For this sampled subset (of N=142k), we compare three mixtures:
\begin{enumerate}
    \item \textbf{1 Language (source)}: This is the monolingual baseline where the SFT data contains code exclusively in the source language of downstream RL training. This represents the standard setting where one relies solely on the high-resource language available for both SFT and RL. We ensure that all 3,111 questions in our dataset are included, uniformly sampling solutions within each question.

    \item \textbf{8 Languages (parallel)}: This is our proposed method. We distribute the 142k instances equally across the 8 languages and all 3,111 questions. \emph{All} questions appear in \emph{all} languages. We use Python, C, C++, Java, C\#, JavaScript, Bash, and Lua, the same as in \S\ref{sec:eval}.

    \item \textbf{8 Languages (non-parallel)}:
    To isolate the effect of parallel programs from simply seeing more PLs during SFT, we construct a PL-disjoint mixture.
    We ensure that no question appears in more than one language, removing the parallel alignment signal.
    The solution distribution across languages is still uniform.
\end{enumerate}
These mixtures have the same diversity as they cover the same number of questions (N=3,111) and the same number of solutions (N=142k). They differ in the number of solutions per question and per language.\footnote{It may also be tempting to compare to the original Tulu coding mixture, but this would be confounded by distinct dataset properties, e.g., data complexity, the PLs covered, etc.}

To put the transfer performance into perspective, we also consider an oracle: \textbf{1 Language (target)}. It is analogous to 1 Language (source) but the data is exclusively in the target language of the downstream transfer. In the subsequent RL stage, this SFT model is always paired with target-language RL training data, representing the ideal "data-rich" scenario where transfer is unnecessary.

\ifonecolumn
\else
\begin{figure*}[t!]
    \begin{subfigure}[t]{\textwidth}
        \centering
        \includegraphics[width=\textwidth]{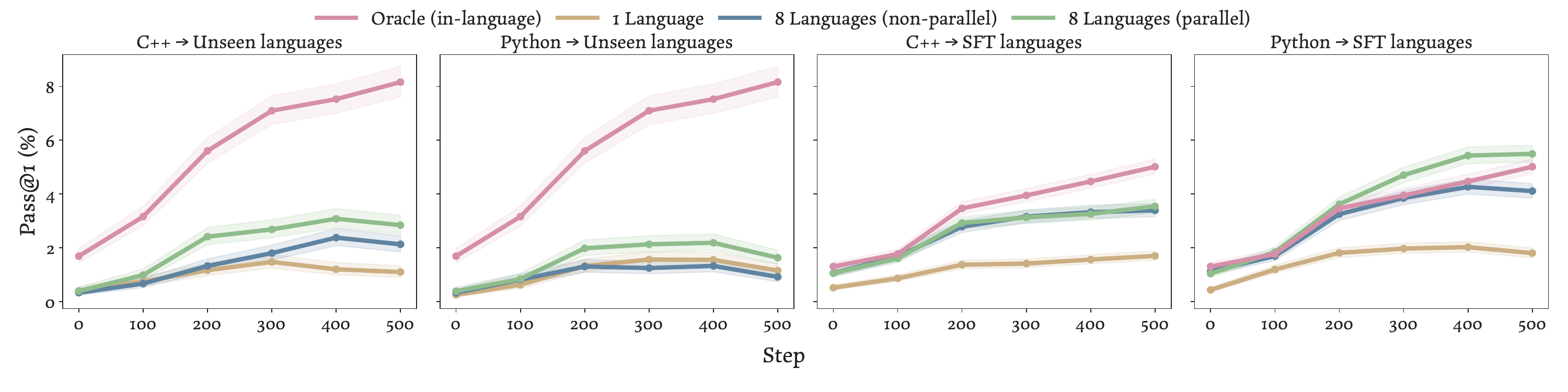}
        \vspace{-7mm}
        \caption{Code generation.}
        \vspace{3mm}
        \label{fig:results-generation}
    \end{subfigure}
    \\
    \begin{subfigure}[t]{\textwidth}
        \centering
        \includegraphics[width=\textwidth]{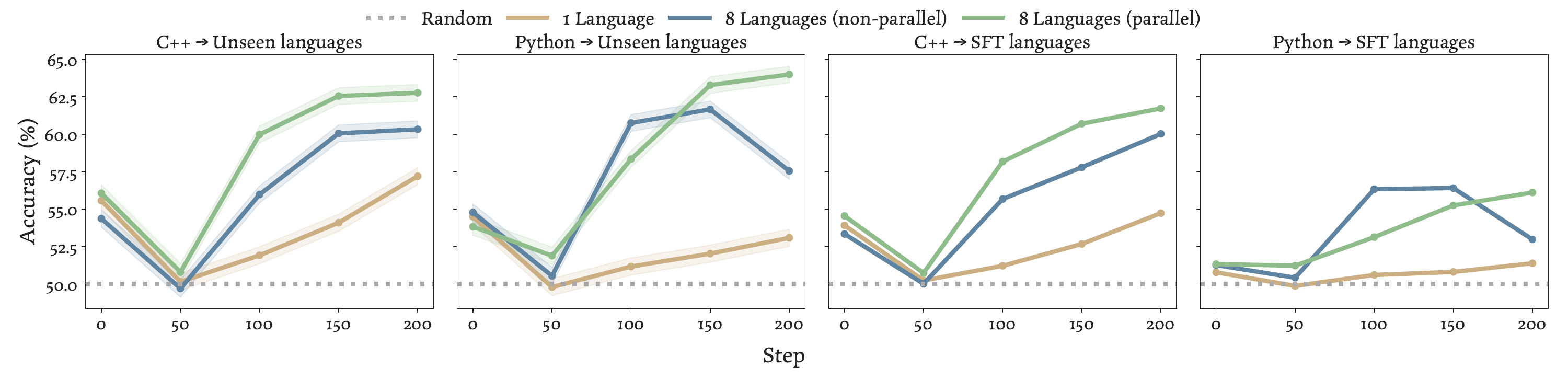}
        \vspace{-7mm}
        \caption{Code validation.}
        \label{fig:results-validation}
    \end{subfigure}
    \caption{\textbf{Parallel-SFT improves cross-PL RL transfer.} 
    We report performance across two source languages and two coding tasks.
    We report the (average) transfer effect to 3 unseen languages (left) and the SFT languages (right; 7 PLs for code generation \emph{excluding the source}; 5 for code validation (see Footnote~\ref{fn:validation-sft-pls})). Shaded regions denote 95\% confidence intervals.
    Parallel-SFT yields the best transferability, outperforming the baselines and, in some cases, surpassing the target-language oracle. Non-parallel multi-PL training sometimes is beneficial too.}
\end{figure*}

\fi

\subsection{Training Setup} \label{sec:setup-rl}

We evaluate the transfer effect from a source PL to another target PL.
For the source, we consider C++ and Python, two high-resource languages.
For the target PL, we consider Go, PHP, and Ruby, the same as in \S\ref{sec:eval}, all of which are unseen during SFT.
We also consider the other SFT languages as target languages whenever appropriate.

We initialize from the Llama-3.1-8B base model\footnote{See Footnote~\ref{fn:why-not-qwen} for our rationale.} and train 3 SFT models on the mixtures in \S\ref{sec:sft-dataset},\footnote{Note that we have two 1 Language (source) models corresponding to the two source languages.} as well as the oracle.
For the oracle, we continue from the 1 Language (target) SFT model and further RL-train it with target PL data to get the ``data-rich'' in-language training performance.

We consider the two coding tasks in \S\ref{sec:coding-tasks}.
For code generation, we use CodeForces, following \S\ref{sec:eval}.
For code validation, we use the CodeForces submissions dataset~\footnote{\url{https://huggingface.co/datasets/open-r1/codeforces-submissions}} that consists of human submissions to the CodeForces website. It contains submissions in multiple PLs, including all of our 3 target languages.
See dataset statistics and additional training details in \S\ref{sec:rl-dataset-stats}.

During evaluation, we sample 8 times per question for both tasks. Pass@8 for code generation requires all 8 samples. For pass@1 as well as boolean accuracy for code validation,\footnote{Though accuracy is the metric, we do not train a classification model because LMs can develop chains-of-thought.} we treat them as independent samples to reduce estimation variance.

\section{Results} \label{sec:results}

\ifonecolumn
\begin{figure*}[t!]
    \begin{subfigure}[t]{\textwidth}
        \centering
        \includegraphics[width=\textwidth]{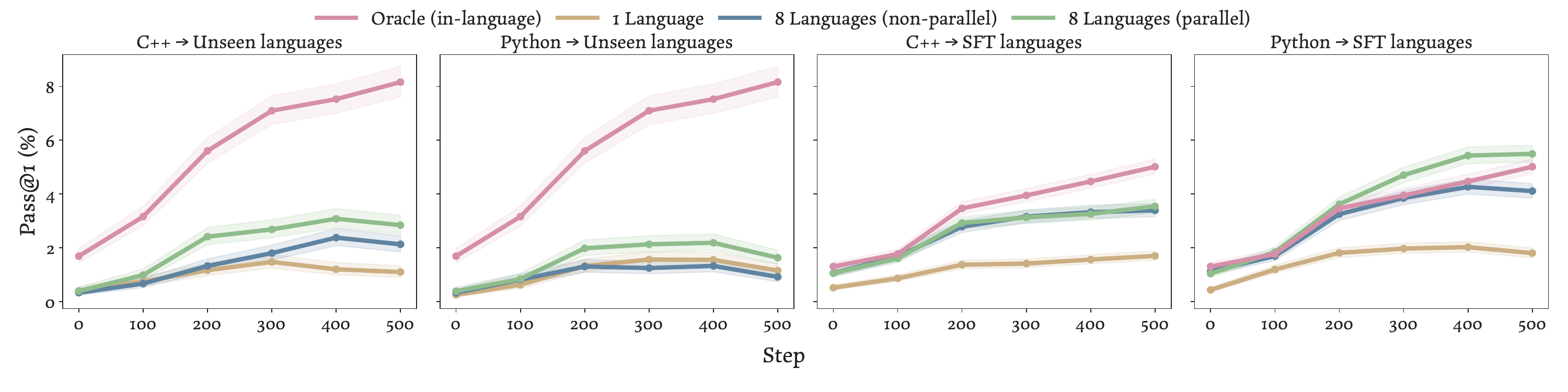}
        \vspace{-7mm}
        \caption{Code generation.}
        \vspace{3mm}
        \label{fig:results-generation}
    \end{subfigure}
    \\
    \begin{subfigure}[t]{\textwidth}
        \centering
        \includegraphics[width=\textwidth]{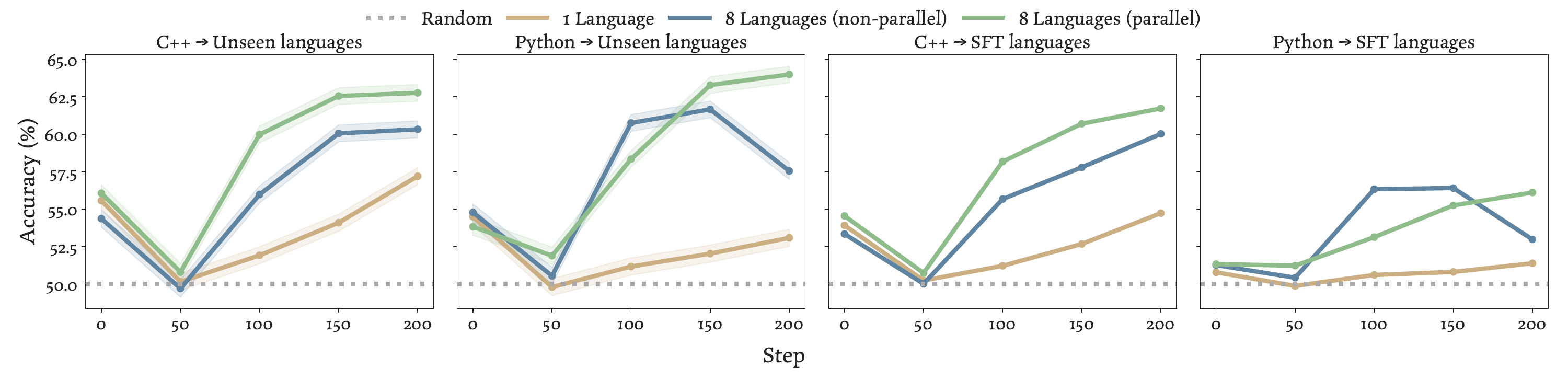}
        \vspace{-7mm}
        \caption{Code validation.}
        \label{fig:results-validation}
    \end{subfigure}
    \caption{\textbf{Parallel-SFT improves cross-PL RL transfer.} 
    We report performance across two source languages and two coding tasks.
    We report the (average) transfer effect to 3 unseen languages (left) and the SFT languages (right; 7 PLs for code generation \emph{excluding the source}; 5 for code validation (see Footnote~\ref{fn:validation-sft-pls})). Shaded regions denote 95\% confidence intervals.
    Parallel-SFT yields the best transferability, outperforming the baselines and, in some cases, surpassing the target-language oracle. Non-parallel multi-PL training sometimes is beneficial too.}
\end{figure*}

\fi

Figure~\ref{fig:results-generation} presents the transfer performance for code generation.
On the left, we visualize the zero-shot transfer results to the 3 target PLs unseen during SFT.
When C++ is the source language, simple (non-parallel) multi-PL training already offers generalization benefits over the monolingual baseline.
Theoretically, this is consistent with prior literature that posited multi-task training as a form of meta-learning that facilitates better downstream adaptation~\citep{wang2021bridging}. Empirically, this aligns with the multilingual literature where more training languages enable better generalization~\citepia{arivazhagan2019massivelymultilingualneuralmachine,aharoni-etal-2019-massively}.
Parallel-SFT enables further improvements and leads to the best transfer performance throughout RL.

On the right, we show the performance when the target PL is in the 8-Language SFT mixture.\footnote{Here, comparing the 8 Languages models with the 1 Language model is no longer fair because the latter does not see these languages during SFT, unlike the former.}
Remarkably, when Python is the source PL, Parallel-SFT not only outperforms the baselines but even the in-language oracle.\footnote{We conjecture that this may be due to models learning more effectively on higher-resource PLs, analogous to multilingual models learning more effectively in English. We leave further studies on this phenomenon to future work.}
This supports our hypothesis that Parallel-SFT creates a high-quality ``generalist'' initialization that allows models to leverage abundant high-quality source data effectively, compared to SFT- and RL-training on the target PL alone, which may suffer from noise.
Additional results measured in pass@8 (\S\ref{sec:additional-results}) show similar trends.

\ifonecolumn
\else
\begin{figure*}[t!]
    \centering
    \begin{subfigure}[t]{0.485\textwidth}
        \centering
        \includegraphics[width=\textwidth]{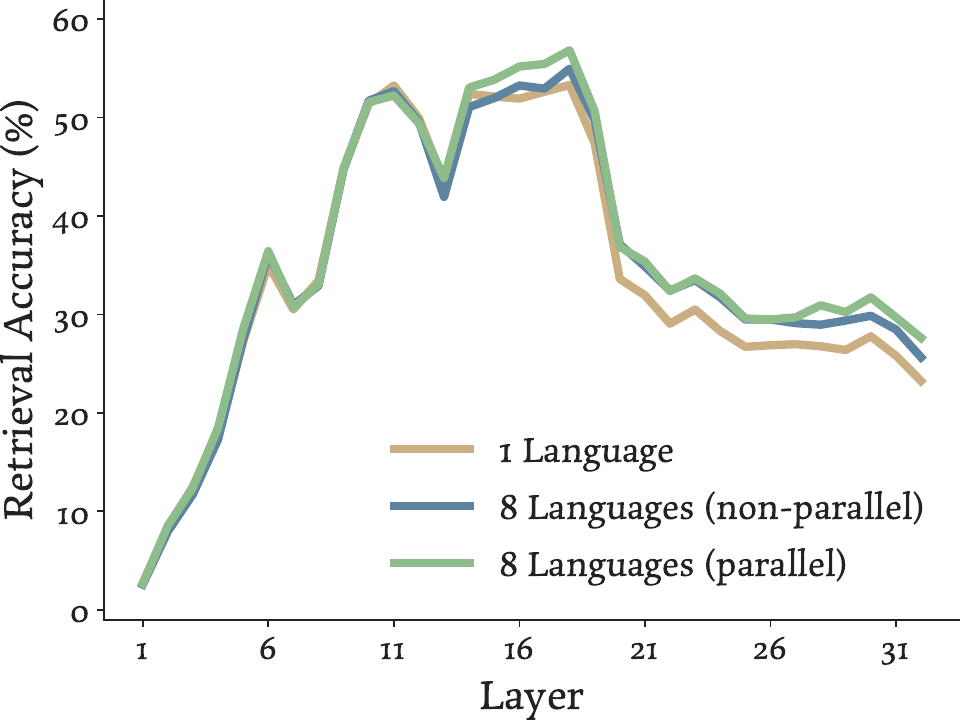}
        \caption{Retrieval accuracy}
    \end{subfigure}%
    ~
    \begin{subfigure}[t]{0.485\textwidth}
        \centering
        \includegraphics[width=\textwidth]{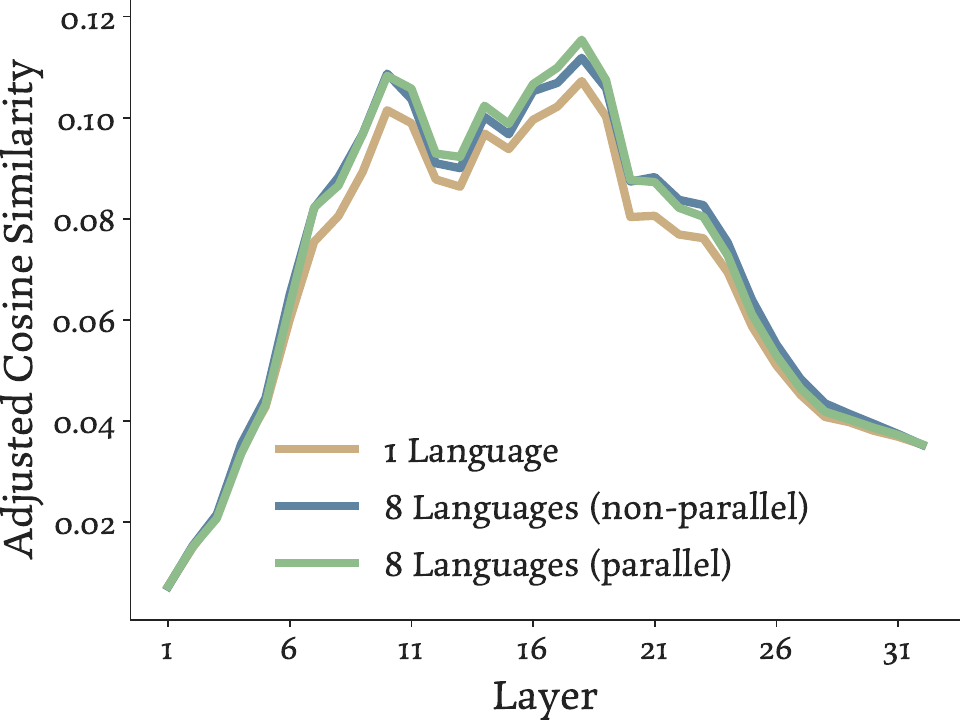}
        \caption{Adjusted cosine similarity}
    \end{subfigure}%
    \caption{\textbf{Parallel-SFT improves representation generality.}
    We measure the cross-PL alignment of program representations for unseen languages across model layers.
    Non-parallel multi-PL training induces more aligned parallel program representations, while Parallel-SFT further improves them. The inverted U-shape suggests that alignment peaks in the middle layers, where semantic reasoning occurs.}
    \label{fig:analysis}
\end{figure*}

\fi

While the code generation results validate the effectiveness of Parallel-SFT, we acknowledge that zero-shot transfer
is primarily a scientific question for this task, but not a practical challenge.
This is because its training instances are PL-agnostic, with only a natural language prompt and stdin-stdout tests. When we ``train in a source language,'' we instruct the model to generate programs in that language and execute the output in that language. Therefore, while the improved transferability from Parallel-SFT is still scientifically meaningful, we also consider the code validation task. In code validation, the code program in the input \emph{is} PL-dependent, making zero-shot transfer a real practical problem. Figure~\ref{fig:results-validation} summarizes the results.\footnote{Because of the realistic data constraint, we no longer have sufficient target-PL RL data for the oracle results for code validation. Additionally, the RL dataset lacks data in Bash and Lua, so when we consider the SFT languages as the target PLs, we only average over 5 languages (8 training languages minus the source PL itself as well as Bash and Lua).\label{fn:validation-sft-pls}} Overall, we observe similar trends to code generation, corroborating the utility of Parallel-SFT.

Furthermore, in \S\ref{sec:in-language-performance}, we also confirm that Parallel-SFT, while achieving improved cross-language generalizability, does not sacrifice in-language performance.

\section{Analysis}

\ifonecolumn
\begin{figure*}[t!]
    \centering
    \begin{subfigure}[t]{0.485\textwidth}
        \centering
        \includegraphics[width=\textwidth]{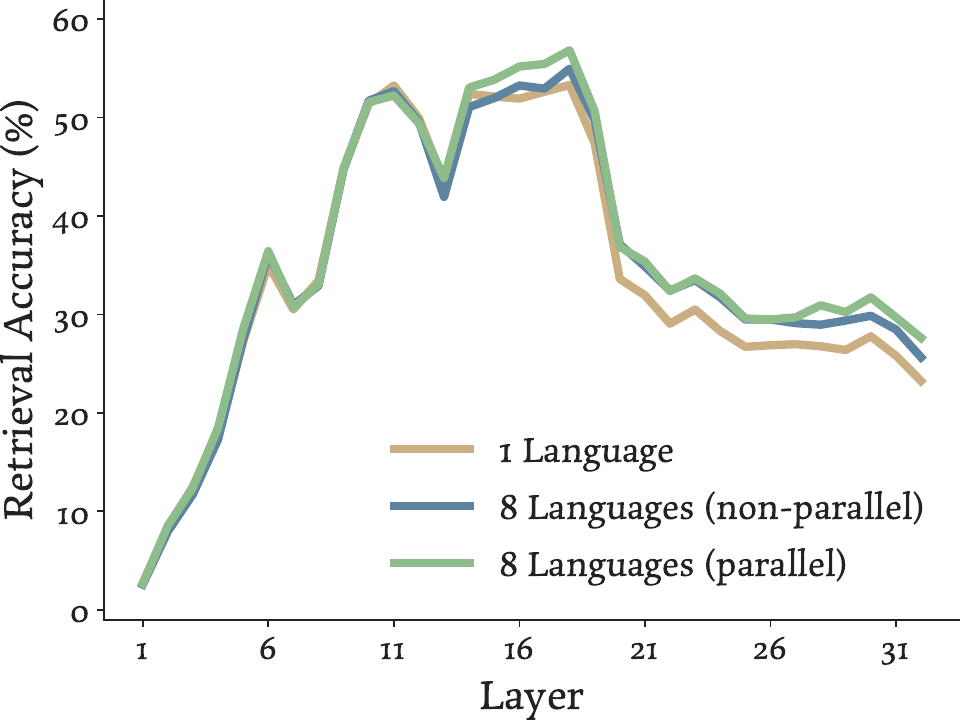}
        \caption{Retrieval accuracy}
    \end{subfigure}%
    ~
    \begin{subfigure}[t]{0.485\textwidth}
        \centering
        \includegraphics[width=\textwidth]{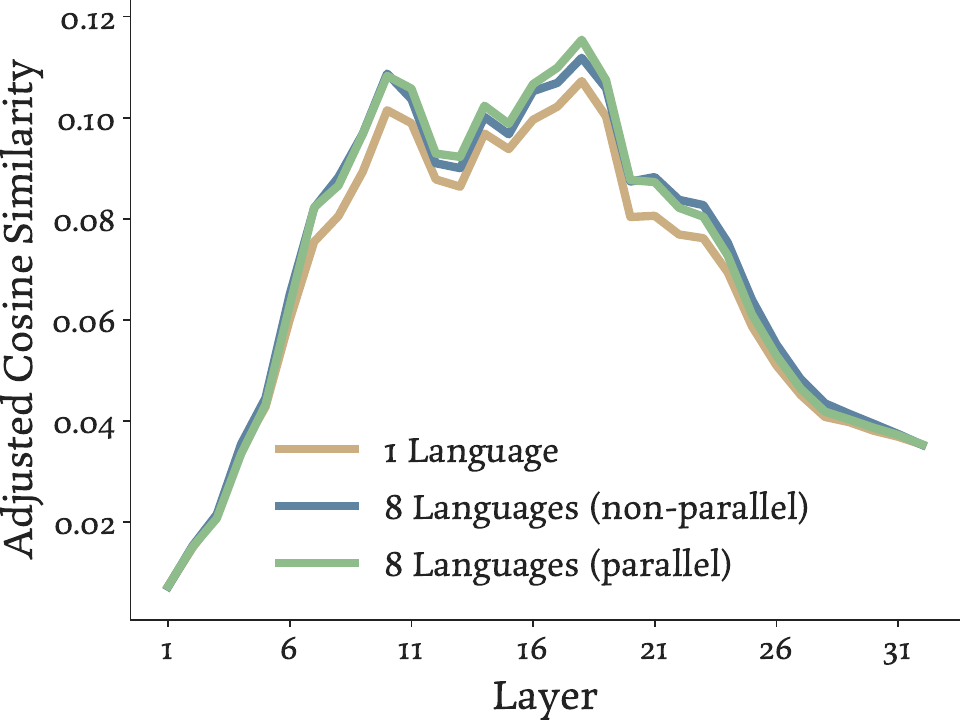}
        \caption{Adjusted cosine similarity}
    \end{subfigure}%
    \caption{\textbf{Parallel-SFT improves representation generality.}
    We measure the cross-PL alignment of program representations for unseen languages across model layers.
    Non-parallel multi-PL training induces more aligned parallel program representations, while Parallel-SFT further improves them. The inverted U-shape suggests that alignment peaks in the middle layers, where semantic reasoning occurs.}
    \label{fig:analysis}
\end{figure*}

\fi

Inspired by the multilingual NLP literature~(\S\ref{sec:bg-nl}), we investigate whether the improved transferability correlates with more PL-general representations.
Specifically,
we check if Parallel-SFT promotes the representation similarity between standalone parallel programs (i.e., without contexts), as is standard in multilingual research~\citepia{wu2025the,shen-etal-2025-unaligned,gao-etal-2025-understanding}.

We construct a held-out dataset with 312 code programs
from our SFT distribution,
each from a different question.
None of the questions has been observed in SFT.
We consider translations of these programs in 3 languages unseen during SFT---Go, PHP, and Ruby---obtained from the synthetic translation procedure in \S\ref{sec:parallel-data-construction}.
This yields a list of parallel programs $\{(c_i^{\text{Go}}, c_i^{\text{PHP}}, c_i^{\text{Ruby}})\}_{i=1}^{N}$ where $N=312$.

For each SFT model, we derive program representations, separately at each layer $\ell$, e.g. $\mathbf{r}_{i,\ell}^{\text{Go}}$. We compute the program representations via echo embedding~\citep{springer2025repetition} that repeats the program twice in a lightweight template which empirically produces strong sequence representations.\footnote{The intuition is that autoregressive models do not attend to the future, but future information is helpful for a good representation. By repeating the sequence twice, the early tokens in the second occurrence can attend to future tokens in the first occurrence. 
We lightly adapt the template in \citet{springer2025repetition}: ``Rewrite the following code: \{x\}. The rewritten code: \{x\}'' where ``\{x\}'' is substituted with the code program.
We mean-pool the token representations in the second ``\{x\}''.}

Representation similarity is most straightforwardly computable by cosine similarity. It, however, can be misleading due to the \emph{anisotropy} of the representation space~\citep{ethayarajh-2019-contextual}. That is, it is possible that a model simply utilizes a small subset of its representation space for code, embedding \emph{all} programs closer together, not just parallel programs.
We account for this in two robust metrics:
\begin{enumerate}
    \item \textbf{Retrieval accuracy} measures whether a program and its translation have closer representations than all other non-corresponding programs~\citepia{artetxe-schwenk-2019-massively,feng-etal-2022-language,shen-etal-2025-unaligned}. Formally, e.g., we measure the accuracy $\cos(\mathbf{r}_{i,\ell}^{\text{Go}}, \mathbf{r}_{i,\ell}^{\text{PHP}}) > \max_{j\ne i}{\cos(\mathbf{r}_{i,\ell}^{\text{Go}}, \mathbf{r}_{j,\ell}^{\text{PHP}})}$.

    \item \textbf{Adjusted cosine similarity.} Following \citet{wu2025the}, we correct for baseline anisotropy by subtracting the cosine similarities between non-parallel pairs. Formally, e.g., we measure $\cos(\mathbf{r}_{i,\ell}^{\text{Go}}, \mathbf{r}_{i,\ell}^{\text{PHP}}) - \E_{j\ne i}[\cos(\mathbf{r}_{i,\ell}^{\text{Go}}, \mathbf{r}_{j,\ell}^{\text{PHP}})]$. We estimate the second term with a single random sample.
\end{enumerate}

We show the results in Figure~\ref{fig:analysis}. We observe that simple (non-parallel) multi-PL training already improves program alignment over the monolingual baseline, and explicit parallel program training further promotes it.
This confirms that Parallel-SFT induces more functionality-centric representations.

Furthermore, interestingly, the similarity measures both follow an inverted U-shape, higher in intermediate layers and lower on the two ends. This is a familiar trend in multilingual studies~\citepia{pires-etal-2019-multilingual,wu-dredze-2019-beto,conneau-etal-2020-emerging,wu2025the}.
The initial/final layers need to map from/to the surface form of PLs and are thus more specialized to the syntax of individual PLs.
The middle layers, on the other hand, serve as a ``semantic hub'' and perform more abstract reasoning~\citep{wendler-etal-2024-llamas,wu2025the}.
Parallel-SFT accentuates this trend: for example, its adjusted cosine similarities are similar to the baseline models on the two ends, but notably higher in the middle layers, where more PL-independent reasoning occurs.

\section{Discussion and Related Work}

Multilingual representation learning is a foundational problem in NLP, from traditional word embeddings~\citepia{mikolov2013exploiting,al-rfou-etal-2013-polyglot,ammar2016massivelymultilingualwordembeddings} to later sequence models~\citepia{schwenk-douze-2017-learning,EspaaBonet2017AnEA,eriguchi2018zeroshotcrosslingualclassificationusing}.
These models achieve remarkable zero-shot cross-lingual generalizability, both on early classification tasks such as entailment~\citepia{huang-etal-2019-unicoder,xlm,conneau-etal-2020-unsupervised,hu2020xtreme} and more recently on tasks in the LM post-training pipeline, such as SFT and reward modeling~\citepia{wu-etal-2024-reuse,shaham-etal-2024-multilingual,chen-etal-2024-monolingual,shimabucoro2025post}.

This success is typically driven by two factors: \textbf{structural isomorphism} and \textbf{explicit alignment signals}.
Early approaches rely on the latter, providing explicit supervised signals such as bilingual lexicons~\citepia{mikolov2013exploiting,kocisky-etal-2014-learning,faruqui-dyer-2014-improving,ammar2016massivelymultilingualwordembeddings} or parallel texts~\citepia{schwenk-douze-2017-learning,artetxe-schwenk-2019-massively,xlm,huang-etal-2019-unicoder}.
Later work showed that representation alignment could also emerge from purely unsupervised signals, due to structural isomorphism alone~\citep{lample2018word,lample-etal-2018-phrase}, although explicit parallel data is still helpful~\citep{reid-artetxe-2023-role,shen-etal-2025-unaligned}.
However, we argue that these factors are often absent for programming languages.

First, natural languages have distributionally similar lexicons, where most words have translations in other languages.
However, PLs' lexical structures usually differ due to heterogeneous paradigms. For instance, non-object-oriented languages lack keywords related to classes; some PLs rely on specific control flow structures like \texttt{switch} statements or lazy returns (\texttt{yield}) that others lack; and mechanisms such as garbage collection and concurrency can be explicitly specified via certain keywords in some PLs but are implicit in others.
Even for shared concepts, usage distributions diverge vastly: while the \texttt{boolean} type is universal, it may be syntactically required in Java (\texttt{boolean x = true;}) but implicitly typed in Python (\texttt{x = True}).

Beyond structure similarities, parallel natural language sentences provide explicit alignment signals, whether deliberately leveraged in pretrained objectives or organically emerging in unsupervised corpora.
However, as we discussed in \S\ref{sec:method}, they are often lacking for programming languages too.
Without these explicit signals or the implicit structural isomorphism of natural languages, zero-shot transfer in code becomes challenging.

Prior work has attempted to bridge this gap through either data-centric or structure-centric methods.
Data-centric approaches attempt to scale up multi-PL datasets, but existing resources are typically scarce beyond small-scale evaluation benchmarks~\citepia{khan-etal-2024-xcodeeval,chai2025mceval,xu-etal-2025-cruxeval}.
Others have synthetically generated data in low-resource PLs via translation~\citep{multiplt}, a practice termed ``translate-train'' in the multilingual NLP nomenclature~\citep{conneau-etal-2018-xnli}.
Alternatively, structure-centric approaches enforce alignment by leveraging language-agnostic intermediate forms such as Abstract Syntax Trees~\citepia{guo-etal-2022-unixcoder,chen-etal-2023-pass,gong2024astt} and Intermediate Representations~\citep{10123512,szafraniec2023code,paul-etal-2024-ircoder}.

Our approach diverges from these methods.
We augment data via synthetic translation, similar to translate-train methods, but with more flexibility afforded by zero-shot cross-PL transfer rather than requiring task- and language-specific translations.
And unlike structure-centric methods, we do not need parsers or compilers, allowing the model flexibility to learn necessary structures from raw text---analogous to the shift in NLP away from explicit syntactic and semantic structures.
Concurrent work~\citep{yang2025scalinglawscodeprogramming} also leverages parallel programs, but focuses on pretraining scaling laws.

\section{Conclusion}

In this work, we identified unsatisfactory zero-shot cross-PL transferability for Llama-3.1.
To address this, we proposed Parallel-SFT that uses parallel programs to train a more generalizable SFT model prior to RL.
We demonstrated that our method improves transferability across PLs, and our analysis confirms that Parallel-SFT induces a more PL-general representation space through grounding code with execution equivalence. This kind of semantic understanding of code is crucial for many application including coding agents and code retrieval~\citep{gu2025tasks}, and we hope that Parallel-SFT inspires future work to train more general and transferrable code models.

\section*{Limitations}

While Parallel-SFT demonstrates improvements in cross-PL transfer, our exploration was not exhaustive.
We did not perform a search over many design decisions, including SFT data formatting, curricula, or mixing ratios.
Further iterations on these could yield further gains, perhaps additionally leveraging typological similarities between PLs, which would better inform transfer.
Additionally, our experiments focused on fundamental coding tasks. We anticipate that the benefits of Parallel-SFT extend to more complex settings, such as reasoning models and coding agents, but verifying this remains an open question.
In particular, we only conducted experiments on non-reasoning 8B models for cost reasons; we believe our findings are orthogonal to these choices, and the gains from Parallel-SFT will persist in larger reasoning models.
We leave these explorations to future work.

\section*{Acknowledgments}

We thank Alex Gu, Naman Jain, Yuning Mao, Shruti Bhosale, and colleagues at the Meta Superintelligence Labs for discussions and help at various stages of this project.

\clearpage
\newpage
\bibliographystyle{assets/plainnat}
\bibliography{paper}

\clearpage
\newpage
\beginappendix

\section{Cross-Programming-Language Evaluation of Custom-Trained SFT Models} \label{sec:eval-custom-sft}

\begin{figure*}[t!]
    \centering
    \includegraphics[width=\textwidth]{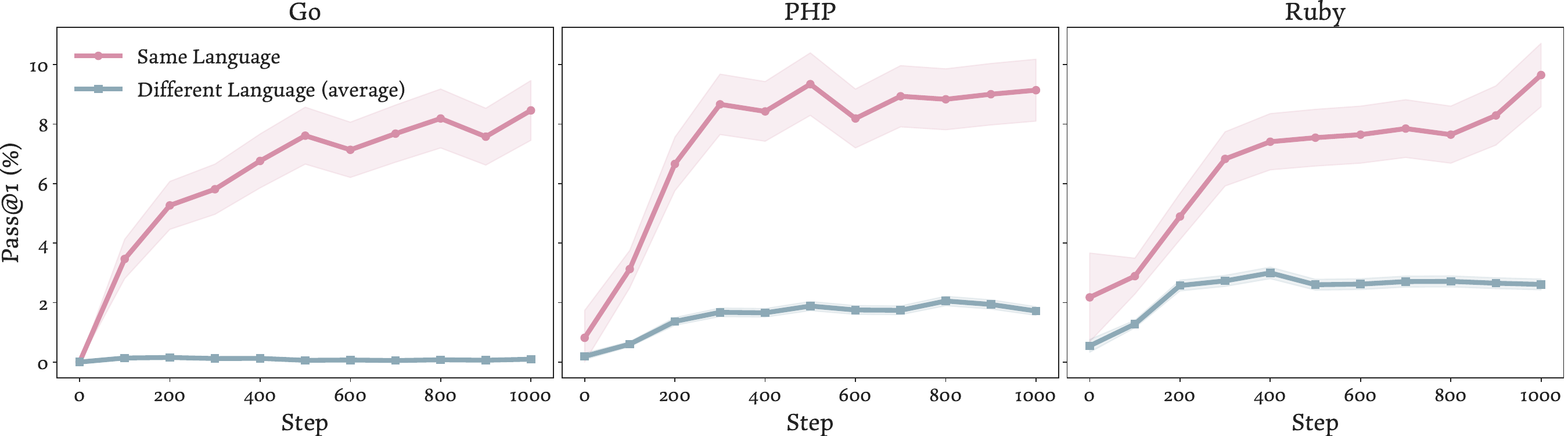}
    \caption{\textbf{Validation of Transfer Failure with Custom SFT.}
    We replicate the experiment from \S\ref{sec:eval} and Figure~\ref{fig:eval} using custom SFT models trained from the Llama-3.1-8B base model, removing the DPO priors.
    We visualize the pass@1 performance on a target language (e.g., Go) during RL training on the same vs. different source languages (averaged over 10 source PLs). Shaded regions indicate 95\% confidence intervals. Consistent with the instruct model results, cross-PL transfer from a different source language leads to limited improvements compared to same-language training.}
    \label{fig:eval-base}
\end{figure*}

In \S\ref{sec:eval}, we observed limited transferability using Llama-3.1-8B-Instruct. It is an SFT + DPO model, and we append a code RL training stage.
While this is consistent with state-of-the-art practices that include multiple stages of policy optimization~\citepia{qwen3,bercovich2025llamanemotronefficientreasoningmodels}, the existing DPO stage may confound transfer effects in the RL stage.
Therefore, we conduct a controlled experiment where we train custom SFT models, initializing from the Llama-3.1-8B \emph{base} model.
We replicate the 1 Language (source) setup from \S\ref{sec:sft-dataset} and Figure~\ref{fig:overview}.
For each of our 11 possible source PLs,
we generate an SFT mixture that contains both the general-purpose Tulu 3 SFT data and
our coding SFT
data exclusively in the \emph{source} language.
The subsequent RL training stage uses the same source PL.
This simulates the setting where we only have access to the source PL data for both SFT and RL.
As shown in Figure~\ref{fig:eval-base}, the results are qualitatively similar to those in \S\ref{sec:eval}: same-language RL consistently leads to performance improvements, but RL-training using a different PL has limited benefits.
This confirms that the transfer failure is fundamental, rather than an artifact of the specific instruct model used.

\section{RL Training Details} \label{sec:rl-dataset-stats}

\paragraph{Code generation.} We use the CodeForces dataset with its existing train/test split. We exclude instances that have custom checker functions as tests and those that do not use stdin-stdout tests. The resulting dataset has 6,617 training instances and 377 test instances. We train with batch size 16.

\paragraph{Code validation.} We use the CodeForces submissions dataset.
We use C++ and Python as our source languages because they have the largest amount of data in this dataset.
For these two PLs, we manually perform a train-test split.
We use batch size 32 and notice that the performance plateaus early and only report performance for 200 steps.
We train with all distinct samples for these 6,400 instances.
For evaluation, we use the remaining instances for each PL.
To prevent class imbalance from skewing the accuracy, we manually balance the correct and incorrect solutions via truncation, on a per-PL basis.
The statistics of the final evaluation instances are:
Python: 19996 samples,
Go: 1502 samples,
PHP: 992 samples,
Ruby: 1150 samples,
Java: 75194 samples,
C++: 19998 samples,
Rust: 534 samples,
Javascript: 2440 samples,
C\#: 8074 samples.

For both settings, we use a learning rate of $5\times10^{-7}$.

\section{Synthetic Translation Prompt} \label{sec:translation-prompt}

We use the below prompt for synthetic translation (\S\ref{sec:parallel-data-construction}). \texttt{\{code\}} is the original Python code snippet in
the coding dataset;
\texttt{\{instruction\}} is the natural language description of the code, from the same source; \texttt{\{lg\_long\}} is the human-readable name of the target programming language (e.g., \texttt{C++}); and \texttt{\{lg\_short\}} is the short name used for Markdown formatting (e.g., \texttt{cpp}).

\begin{mdframed}[backgroundcolor=gray!10, linewidth=0.5pt, roundcorner=5pt]
\small
\ifonecolumn
\begin{verbatim}
Translate the following code from Python to {lg_long}. The functionality should be exactly the same across
all possible inputs.

```python
{code}
```

For reference, the original Python code was generated according to the following request:
'''
{instruction}
'''

Now translate the code into {lg_long} and wrap it in
```{lg_short}
```
Closely follow the stylistic aspects of the original Python code, for example: (1) Respecting the amount
of comment (e.g., function-level/line-level/etc.) and not be over-verbose or under-verbose; (2) Not
writing sample inputs **if** the original Python code doesn't have them; (3) The original Python code
can be directly run with top-level statements. If this is allowed in {lg_long}, follow this behavior; if
this is impossible in {lg_long}, put the top-level statements in a canonical entry point, such as a main
function.
\end{verbatim}
\else
\begin{verbatim}
Translate the following code from Python
to {lg_long}. The functionality should
be exactly the same across all possible
inputs.

```python
{code}
```

For reference, the original Python code
was generated according to the following
request:
'''
{instruction}
'''

Now translate the code into {lg_long}
and wrap it in
```{lg_short}
```
Closely follow the stylistic aspects of
the original Python code, for example:
(1) Respecting the amount of comment
(e.g., function-level/line-level/etc.)
and not be over-verbose or under-verbose;
(2) Not writing sample inputs **if** the
original Python code doesn't have them;
(3) The original Python code can be
directly run with top-level statements.
If this is allowed in {lg_long}, follow
this behavior; if this is impossible in
{lg_long}, put the top-level statements
in a canonical entry point, such as a
main function.
\end{verbatim}
\fi
\end{mdframed}

\section{In-Language Performance of Parallel-SFT} \label{sec:in-language-performance}

We empirically test the extent to which parallel program training degrades in-language performance (i.e., training and testing on the same PL). We performed in-language RL training and evaluation (using Python), starting from both the 8-language non-parallel SFT model and the 8-language Parallel-SFT model. After 500 RL steps, non-parallel training leads to 12.47\% pass@1 (95\% CI: [11.32, 13.71]) and parallel training leads to 12.36\% pass@1 (95\% CI: [11.22, 13.60]). The difference is not statistically significant, confirming that parallel training does not sacrifice in-language performance.

\section{Additional Results} \label{sec:additional-results}

\begin{figure*}[t!]
    \centering
    \includegraphics[width=\textwidth]{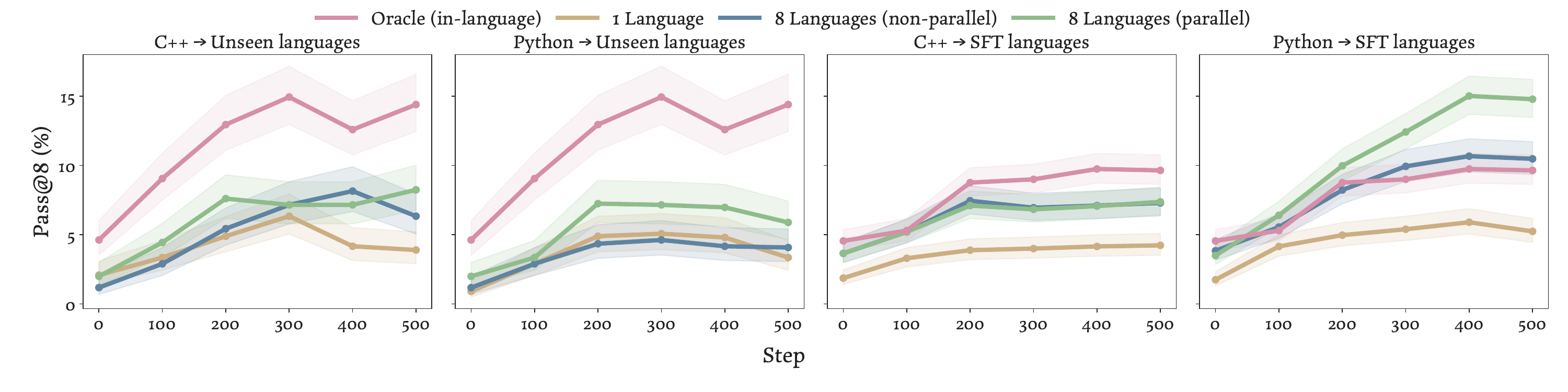}
    \caption{\textbf{Parallel-SFT improves cross-PL RL transfer (pass@8).} 
    We report pass@8 performance across two source languages.
    We report the (average) transfer effect to 3 unseen languages (left) and the SFT languages (right; 7 PLs excluding the source). Shaded regions denote 95\% confidence intervals.
    Consistent with the pass@1 results in Figure~\ref{fig:results-generation}, parallel-SFT yields the best transferability, outperforming the baselines and, in some cases, surpassing the target-language oracle.}
    \label{fig:results-generation-passat8}
\end{figure*}

Figure~\ref{fig:results-generation-passat8} shows the code generation results using the pass@8 metric, complementing the pass@1 results in Figure~\ref{fig:results-generation}. The trends are consistent, corroborating the effectiveness of Parallel-SFT.

\end{document}